\RequirePackage[hyphens]{url}
\documentclass{article}

\usepackage[final]{nips_2018}

\usepackage[utf8]{inputenc} 
\usepackage[T1]{fontenc}    
\usepackage{hyperref}       
\usepackage{booktabs}       
\usepackage{amsfonts}       
\usepackage{nicefrac}       
\usepackage{microtype}      

\usepackage[english]{babel}
\usepackage{blindtext}
\usepackage{multirow}
\usepackage{soul}
\usepackage{amsmath}
\usepackage{todonotes}
\usepackage{xcolor}
\usepackage{enumitem}
\usepackage[linesnumbered,ruled,vlined]{algorithm2e}
\usepackage{subfig}

\title{Interpretable Credit Application Predictions With
Counterfactual Explanations}

\newcommand*{\affaddr}[1]{#1} 

\newcommand*{\email}[1]{\texttt{#1}}

\author{
 Rory Mc Grath\textsuperscript{1}, Luca Costabello\textsuperscript{1}, Chan Le Van\textsuperscript{1}, \\ 
 \textbf{Paul Sweeney\textsuperscript{2}, Farbod Kamiab\textsuperscript{2}, Zhao Shen\textsuperscript{2}, Freddy L{\'e}cu{\'e}\textsuperscript{1,3}}\\
 \affaddr{\textsuperscript{1}Accenture Labs} \email{\{rory.m.mc.grath, luca.costabello, chan.v.le.van\}@accenture.com}\\
 \affaddr{\textsuperscript{2}Accenture The Dock} \email{\{paul.p.sweeney, farbod.kamiab, zhao.shen\}@accenture.com} \\
 \affaddr{\textsuperscript{3}Inria} \, \email{freddy.lecue@inria.fr}\\
}

\begin{document}

\maketitle

\begin{abstract}
We predict credit applications with off-the-shelf, interchangeable black-box classifiers and we explain single predictions with counterfactual explanations.
Counterfactual explanations expose the minimal changes required on the input data to obtain a different result e.g., approved vs rejected application.
Despite their effectiveness, counterfactuals are mainly designed for changing an undesired outcome of a prediction i.e. loan rejected. Counterfactuals, however, can be difficult to interpret, especially when a high number of features are involved in the explanation.
Our contribution is two-fold: i) we propose \textit{positive counterfactuals}, i.e. we adapt counterfactual explanations to also explain accepted loan applications, and ii) we propose two weighting strategies to generate more interpretable counterfactuals. Experiments on the HELOC loan applications dataset show that our contribution outperforms the baseline counterfactual generation strategy, by leading to smaller and hence more interpretable counterfactuals.
\end{abstract}

\blindmathtrue

\section{Introduction}\label{sec:intro}
Explaining predictions of black box models is of uttermost importance in the domain of credit risk assessment \cite{bruckner2018regulating}. The problem is even more prominent given the recent right to explanation introduced by the European General Data Protection Regulation~\cite{goodman2016european}, and a must due to regulation in the financial domain. A common approach to explain black box predictions focuses on generating local approximations of decisions. If $f$ is a machine learning model taking the features $X$ and mapping them to targets $Y$, then the goal is to find a subdomain of the feature variables and over that domain approximate $f \sim g$, where $g$ is an interpretable and easy to understand function. There has been recent interest in model-agnostic methods of explainability. These methods look to create an \textit{explainer} that should be able to explain any model treating the underlying model as a black box. \cite{LIME}. 

This paper focuses on \textit{Counterfactual Explanations}~\cite{main_paper}, one of these model-agnostic methods. A counterfactual explanation may justify a rejected loan application as follows: 

\begin{center}
\textit{Your application was denied because your annual income is \$30,000 and your current balance is \$200. If your income had instead been \$35,000 and your current balance had been \$400 and all other values remained constant, your application would have been approved}. 
\end{center}

The explanation describes the required minimum change in inputs to flip the decision of the black box classifier. Note that the latter remains a black box: it is only through changing inputs and outputs that an explanation is obtained. 

\begin{figure}
\begin{center}
	\includegraphics[width=.9\columnwidth]{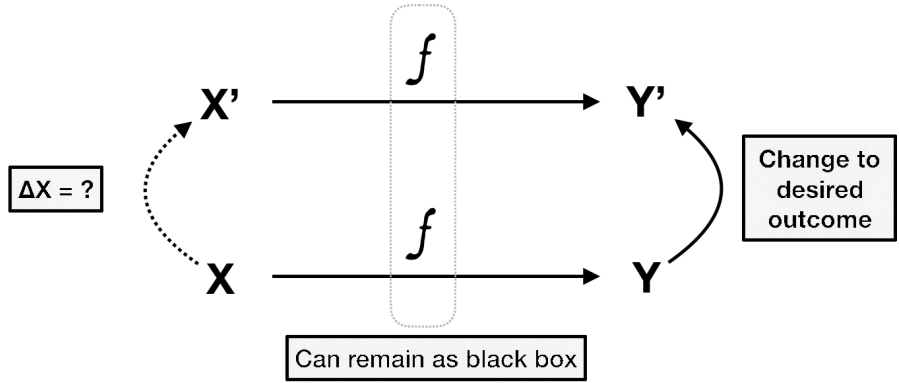}
	\caption{Explaining black box predictions with counterfactuals}
	\label{fig:explainable_2}
	\end{center}
\end{figure}

Despite their effectiveness, two problems arise: on one hand counterfactuals are inherently designed to describe what it takes to flip the decision of a classifier, hence they poorly address the case in which the decision was satisfactory from an end user perspective (e.g. loan approved). 
Problems also arise when assessing the interpretability of counterfactuals: the generated counterfactuals often suggest to change a high number of features, therefore leading to less intelligible explanations. For example, counterfactuals generation strategies do not take into account the \textit{importance} of the dataset features, thus underestimating or overestimating certain dimensions. 
This problem is of particular importance given that it has been showed that human short-term memory is unable to retain a large number of information units~\cite{vogel2001storage,alvarez2004capacity}.
This remains as problematic in the finance domains of loan approval when data scientis and regulators are in the loop.

We predict loan applications with off-the-shelf, interchangeable black-box estimators, and we explain their predictions with counterfactual explanations. To overcome the aforementioned problems, we present the following contribution:

\begin{itemize}  
\item \textbf{Positive Counterfactuals}: in case of a desired outcome, we interpret counterfactuals as a \textit{safety margin}, i.e. a tolerance from the decision boundary. Such counterfactual explanations for positive predictions answer the question \textit{"How much was I accepted by?"}.
\item \textbf{Weighted Counterfactuals}: inspired by Huysmans et al.~\cite{huysmans2011empirical}, we use the size of explanations as a proxy to measure their interpretability. To obtain more compact (and hence more intelligible) counterfactuals we introduce weights in the their generation strategy. We propose two weighting strategies: one based on global feature importance, the other based on nearest neighbours.
\end{itemize}

We experiment on a credit application dataset and show that our weighted counterfactuals generation strategies lead to smaller counterfactuals (i.e. counterfactuals that suggest to change a smaller number of features), thus delivering more interpretable explanations.

\section{Related Work}\label{sec:relatedwork}

\textbf{Local Interpretability}
A number of works focus on explaining \textit{single predictions} of machine learning models, rather than the model as a whole. This task is also known as \textit{local interpretability}. 
White box models come with local explanations by design: traditional transparent design approaches include decision trees and rule extraction \cite{Guidotti:2018:SME:3271482.3236009,molnar}.
In that respect some works such as \cite{DBLP:conf/nips/CravenS95} built surrogate models by interfacing complex models such as deep neural networks with more interpertable models such as decision trees. The authors aim at mimicing the behaviour of a complex model with a much simpler model for interpretability purpose.
Other approaches are instead model-agnostic, and also address explanations of predictions of black box models. LIME generates local explanations from randomly generated neighbours of a record. Features are weighted according to distances from the record.\cite{ribeiro2016model}. SHAP is another approach based on feature importance for each record \cite{NIPS2017_7062}.
Other recent lines of research rely on \textit{example-based explanations}: Prototype~\cite{bien2011prototype} and Criticism \cite{kim2016examples} Selection are two recent examples of this. Prototypes are tuples representative of the dataset, whereas criticisms are examples which are not well-explained by prototypes. Adversarial examples~\cite{kurakin2016adversarial} are another example-based approach, but they are designed to flip the decision of a black-box predictor rather than explaining it. Counterfactual explanations are also an example-based strategy, but unlike adversarial examples, they inform on how a record features must change to radically influence the outcome of the prediction.

\textbf{Interpretable Credit Risk Prediction}
Providing a comprehensive review of more than 20 years of research in credit risk prediction models is out of the scope of this paper. The survey by Lyn et al. \cite{thomas2017credit} gives a comprehensive and up-to-date overview). Huang et al. \cite{huang2004credit} briefly mention an explanation of predictive models for credit rating, but their survey limits to ranking features by importance with variance analysis. A more recent survey by Louzada et al. focuses on predictive power\cite{louzada2016classification} only. Instead, we list works that consider interpretability as a first-class citizen.
A number of works rely on white box machine learning pipeline, mostly using decision trees and rules inference:
Khandani et al. \cite{khandani2010consumer} propose a pipeline to predict consumer credit risk with manual feature engineering and decision trees. The latter being an explainable model, we can consider this work as a rather interpretable approach.
Florez-Lopez et al. adopt an ensemble of decision trees and explain predictions with rules \cite{Lpez2015EnhancingAA}. Predictive power is encouraging, but there is no comparison against neural architectures. Martens at al. combine rule extraction with SVMs \cite{martens2007comprehensible}.
Obermann et al. compare decision tree performance to grey and black boxes approaches on an insolvency prediction scenario \cite{obermann2015demonstrating,obermann2016interpretable}.
Other white-box approaches include Markov models for discrimination \cite{volkov2017incorporating} and rule inference \cite{xu2017improved}. 
Black box approaches show the most promising predictive power, to the detriment of interpretability. Danenas et al adopt SVM classifiers~\cite{danenas2015selection}. Addo et al. leverage a number of black-box models, including gradient boosting and deep neural architectures \cite{addo2018credit}. Although they evaluate the predictive power of their models they do not attempt to explain their predictions, either locally or globally. To the best of our knowledge, no work in literature focuses on local interpretability for black box models applied to credit risk prediction.

\section{Preliminaries: Counterfactual Explanations}\label{sec:background} 

A counterfactual explanation describes a generic causal situation in the form:

\begin{quotation}
\textit{Score $y$ was returned because variables $X$ had values $(x_1,x_2...)$ associated with them. If $X$ instead had values $(x'_1,x'_2,...)$, and all other variables had remained constant, score $y'$ would have been returned.}
\end{quotation}
Counterfactuals do not need to know the internal structure or state of model or system (e.g. neural network, logistic regression, support vector machine, etc).
We treat $f$ as a black box that takes the feature vector $x$ and generates the outcome $y$, and we determine what is the \textit{closest} $x'$ to $x$ that would change the outcome of the model from $y$ to the desired $y'$ (Figure~\ref{fig:explainable_2}).
When generating counterfactuals it is assumed that the model $f$, the feature vector $x$ and the desired output $y'$ are provided. The challenge is finding $x'$, i.e. an hypothetical input vector which falls close to $x$ but also for which $f(x')$ falls sufficiently close to $y'$. 

\textbf{Generating Counterfactuals.} 
We generate counterfactual explanations by calculating the smallest possible change ($\Delta X$) that can be made to the input $X$, such that the outcome flips from y to $y'$. We generate counterfactuals by optimizing the following loss function $\mathcal{L}$, as proposed by Wachter et al.~\cite{main_paper}:

\begin{equation}
\mathcal{L}(x,x',y',\lambda) = \lambda (\hat{f} (x')-y')^2 + d(x,x')
\label{loss_function}
\end{equation}

\begin{equation}
	arg \min_{x'} \max_{\lambda} \mathcal{L}(x,x',y',\lambda)
\end{equation}

where $x$ is the actual input vector, $x'$ is counterfactual vector, $y'$ is the desired output state, $\hat{f}(...)$ is the trained model, $\lambda$ is the balance weight. $\lambda$ balances the counterfactual between obtaining the exact desired output and making the smallest possible changes to the input vector $x$. Larger values for $\lambda$ favor counterfactuals $x'$ which result in a $\hat{f}(x')$ that comes close to the desired output $y'$, while smaller values lead to counterfactuals $x'$ that are very similar to $x$.
The distance metric $d(x,x')$ measures $\Delta{x}$, i.e. the amount of change between $x$ and $x'$. We use the Manhattan distance weighted feature-wise with the inverse median absolute deviation (MAD)~\ref{distance_2}. Such metric is robust to outliers, and introduces sparse solutions where most entries are zero~\cite{main_paper}. Indeed, the ideal counterfactual is one in which only a small number of features change and the majority of them remain constant. The distance metric $d(x,x')$ can be written as: 
\begin{equation}
	d(x, x') = \sum_{j=1}^{p} {|x_j - x'_j| \over MAD_j },
	\label{distance_2}
\end{equation}
\begin{equation}
	MAD_j = median_{i \in \{1, ...,n\}} \Big( \Big| x_{i,j} - median _ { l \in \{1, ..., n\}}   (x_{l,j}) \Big| \Big).
	\label{normalized_formula}
\end{equation}

To generate counterfactuals we adopt the iterative approach described in Algorithm~\ref{algo.gen}. We optimize $\mathcal{L}$ with the Nelder-Mead algorithm, as suggested in~\cite{molnar}.
We constrain the optimisation with a tolerance $\varepsilon$ s.t $|\hat{f}(x')-y'| \leq \varepsilon$. The value for $\varepsilon$ depends on the problem space and is determined by the range and scale of $y$. 
Step 3 iterates over $\lambda$ until the $\varepsilon$ constraint is satisfied. 
A check is performed for a value greater than $\varepsilon$ as increasing $\lambda$ will place more weight on obtaining an $\hat{f}(x')$ closer to the given desired output $y'$.
Once an acceptable value for $\lambda$ is obtained for the given $x$ and $y'$ a set of counterfactuals can be obtained by repeating steps 1 and 2 with the calculated $\lambda$.
Note that we constrain the features manually, since the heuristic in Algorithm~\ref{algo.gen} and the adopted optimization algorithm are designed for unconstrained optimization.

\begin{algorithm}    
    sample a random instance as the initial $x'$ \\
    optimise $L(x,x',y',\lambda)$ with initial $x'$

    \While{$|\hat{f}(x')-y'| > \varepsilon$}{
        increase $\lambda$ by step-size $\alpha$ \\
        optimise $L(x,x',y',\lambda)$ \ref{loss_function} with new $x'$
    }
   	return $x'$
    
    \caption{Counterfactual generation heuristic}
    \label{algo.gen}
\end{algorithm}

\section{Contribution}\label{sec:method}
In this section we describe our two main contributions made towards the explainability of black box machine learning pipelines that predict credit decisions.
Using {\it positive counterfactuals} we explain why a loan was accepted and provide details that help inform an individual when making future financial decisions. 
Next we present weighted counterfactuals that aim at personalizing the counterfactual recommendations that are provided to individuals that received an undesirable outcome (i.e. their loan was denied).

\subsection{Counterfactuals for positive predictions}
In order to explain why applications were accepted, we applied counterfactuals to the scenario where the individual received the desired outcome, i.e {\it positive counterfactuals}.
Here instead of answering the question "Why wasn't I accepted?" we focus on the question of "How much was I accepted by?".
Such approach informs the individual about the features and value ranges that were important for their specific application, thus favouring more informed decisions about potential future financial activities. For example, if the individual is considering an action that may temporarily increase their number of delinquencies then, armed with {\it positive counterfactuals}, they will have a better understanding of the impact on future loan applications.

In the binary classification case we achieve {\it positive counterfactuals} by setting the target $y'$ to be the decision boundary i.e $P(y=1)=0.5$.
This allows us to identify the locally important features that would push the individual to the threshold of being accepted. Another way of viewing this is that these are the features that locally contribute to the desired outcome.

\begin{figure}
\centering

\subfloat[Positive counterfactual explanation]{\includegraphics[width=\columnwidth]{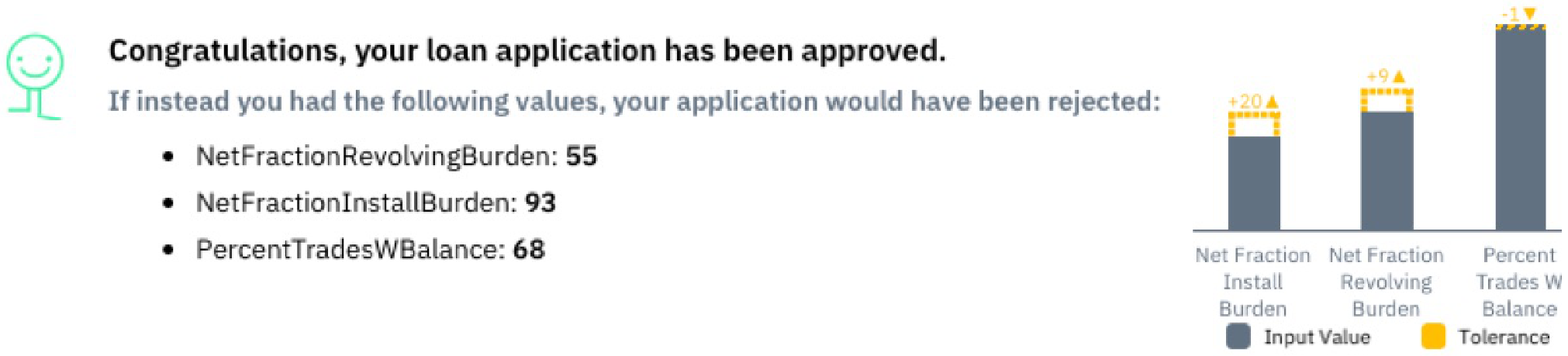}}\\
\subfloat[Counterfactual explanation]{\includegraphics[width=\columnwidth]{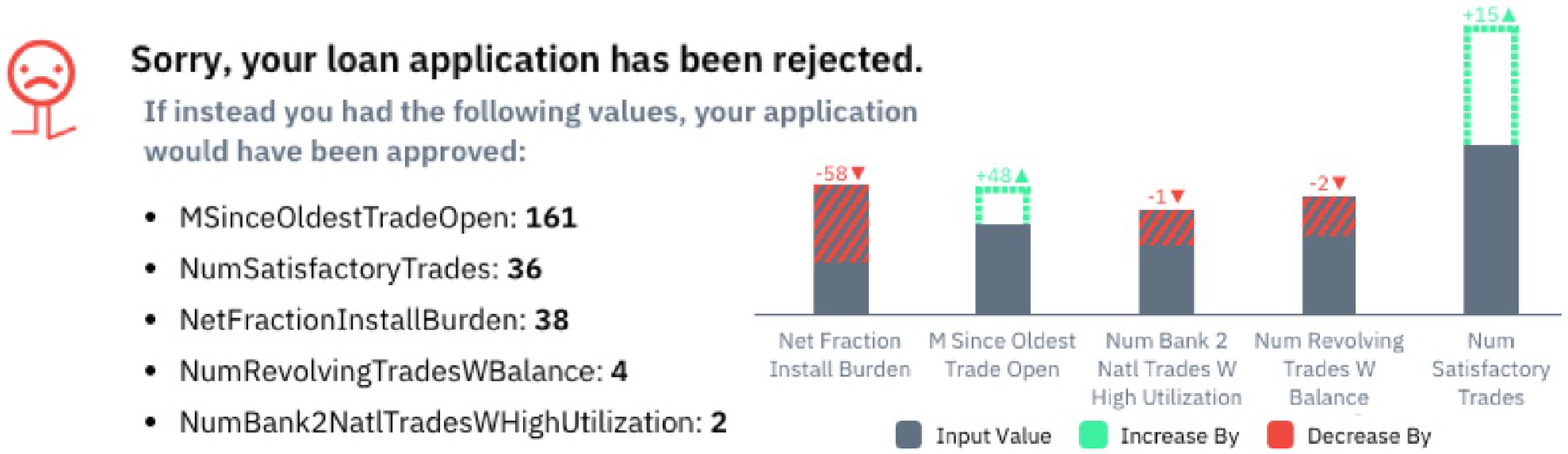}}

\caption{Graphical depictions of a positive (a) and negative (b) counterfactual explanation. Note (a) answers the question \textit{"How much was I accepted by?"} - thus leading to tolerances (highlighted in yellow), whereas (b) explains why the credit application was rejected. In this case the counterfactual explanation suggests how to increase (green, dashed) or decrease (red, striped) each feature.}
\label{fig:examples}
\end{figure}

We present this information to the individual and display it as \textit{tolerance}. In Figure \ref{fig:examples} this is illustrated by a dashed line. Given that future actions do not reduce the indicated features below the dashed line, and all other features remain constant, then the individuals application should remain likely to be approved.

\subsection{Weighted Counterfactuals} 
\label{sec:weighted_counterfactuals}
The general implementation of counterfactuals described in Section~\ref{sec:background} assumes all features are equally important and changing each feature is equally viable.
This, however, is not necessarily the case. For each feature its ability to change and the magnitude of the change may vary on a case by case bases. 
In order to capture this information and create more interpretable actionable recommendations, the generated counterfactuals need to take this into consideration. 
For example some individuals may be able to increase their savings, while others instead may find it easier to reduce their current expenses. 
There are also cases where some features may be fixed or immutable. 
Features like the number of delinquencies in the last six months is historical and fixed. 
Recommending to change these types of features would be of little use.
Our intuition is that promoting highly discriminative features during the generation of counterfactuals leads to more compact, hence better interpretable explanations~\cite{huysmans2011empirical}.

We address these issues by introducing a weight vector $\theta$ to the distance metric defined in Equation \ref{distance_2}. This vector promotes highly discriminative features. 

\begin{equation}
	d_2(x, x') = \sum_{j=1}^{p} {|x_j - x'_j| \over MAD_j }\theta_j,
	\label{weighted_distance}
\end{equation}

We propose two different strategies to generate these weight vectors. The first relies on the global feature importance, the second relies on a Nearest Neighbors approach. The goal is obtaining counterfactuals that suggest a smaller number of changes or focus on values that are relevant to the individual and have historically been shown to vary. 

\textbf{Global feature importance.} We compute global feature importance using analysis of variance (ANOVA F-values) between each feature and the target, and we create a weight vector that promotes highly discriminative features. 
Our goal is obtaining a smaller set of features in the resulting counterfactual recommendation, thus obtaining more compact explanations.

\textbf{K-Nearest Neighbors.} The second approach uses K-Nearest Neighbors to find cases that are close to the individual but have achieved the desired results. 
Looking at the nearest neighbors and aggregating over the relative changes we build a weight vector $\theta$ that captures the locally important features for this individual that have historically been shown to change. 
Here we aim to find counterfactuals containing features that are more actionable by the individual.
By using K-Nearest Neighbors approach these weights can be automatically learned when applied to new problem spaces.

\section{Experiments}\label{sec:eval}

We perform a a binary classification task on a credit application dataset. We train a range of black box models and we explain their predictions with counterfactuals, the goal being explaining the classifier decision to reject or accept a loan application. We perform two separate experiments: first, we carry out a preliminary evaluation of the predictive power of our pipeline. This is not the primary focus of this paper, but it is a required step to gauge the quality of predictions. In a second experiment, we assess the size of counterfactuals generated by our weighted counterfactuals generation.

\subsection{Experimental Settings}

\textbf{Dataset.}
We experiment with the HELOC (Home Equity Line of Credit) credit application dataset. Used in the FICO 2018 xML Challenge\footnote{\url{https://community.fico.com/s/explainable-machine-learning-challenge}}, it includes anonymized credit applications made by real homeowners. We drop highly correlated features and filter duplicate records. After pre-processing we obtain 9,870 records (of which 5,000 positives, i.e. accepted credit applications), and 22 distinct features.

\textbf{Implementation Details.}
Our machine learning pipeline is written in Python 3.6. This includes preprocessing, training, counterfactuals generation, and performance evaluation. We use scikit-learn 0.20 for the black box classifiers\footnote{\url{http://scikit-learn.org/}}. All experiments were run under Ubuntu 16.04 on an Intel Xeon E5-2620 v4 2.10 GHz workstation with 32 GB of system memory.

\subsection{Results}

\textbf{Predictive Power}
As preliminary experiment, we assess the predictive power of four classifiers: logistic regression (LogReg), gradient boosting (GradBoost), support vector machine with linear kernel (SVC), and multi-layer perceptron (MLP). Logistic regression apart, the others fall within the black box category. 
We perform 3-fold, cross-validated grid search model selection over a number of hyperparameters. We adopt balanced class weights for logistic regression, exponential loss for gradient boosting, for each dataset. SVM uses balanced weights, $C=.001$. The neural network uses one hidden layer with 22 units. We use the logistic activation function. Where not specified, we rely on scikit-learn defaults.
Results in Table~\ref{table:results_power} show the predictive power of the best models. Metrics are 3-fold cross-validated.

\begin{table}
  \caption{Predictive power of the adopted black box classifiers. Best results in bold.}
  \label{table:results_power}
  \centering
    \footnotesize
  \begin{tabular}{@{\extracolsep{9pt}}l cc@{}}
      \toprule
              & \multicolumn{2}{c}{\textbf{HELOC}} \\
      \cline{2-3} 

      Model & F1 & Acc  \\
      \midrule

     LogReg
        & \textbf{0.72}      & 0.73 \\

     MLP
        & 0.70      & 0.71 \\

     GradBoost
        & \textbf{0.72}     & \textbf{0.74} \\

     SVC
        & \textbf{0.72}     & 0.73 \\

      \bottomrule
  \end{tabular}
\end{table}

\textbf{Counterfactuals Size}
The preliminarily results of the different weighting strategies as described in Section \ref{sec:weighted_counterfactuals} are presented in Table \ref{table:results_size}.
We experiment with 5,000 loan applications in the dataset: we generate a counterfactual explanation for each of them, and compute the average counterfactuals size. Results show that both weighted strategies bring counterfactuals that have a smaller mean and standard deviation. 
We also observed that in general the average size of the counterfactual recommendations can vary dramatically for the same data given the underlying model.

In general the global feature importance results in features with a lower mean and standard deviation. We obtain explanations which are $11.2\%$ smaller on average using the global feature importance strategy against the baseline.
This is to be expected, as we promote more discriminative features and as a consequence less ancillary features are required.
The benefit in the KNN approach is that the counterfactuals are weighted on the features that are locally important. 
Here we see that while they may not be the best approach they never perform worse than the baseline.
The benefit of the weighting strategies comes with helping the optimization process converge on a local optimum when the underlying space is complex. We look to investigate this claim in future work. 

\begin{table}
  \caption{Average size (i.e. average number of features) of generated counterfactual explanations, for each adopted black box classifier. Smaller counterfactuals mean more interpretable explanations. \textit{Importance}=global feature importance strategy, \textit{KNN}=k-nearest neighbours. KNN uses $k=20$. Best results in bold.}
  \label{table:results_size}
  \centering
    \footnotesize
  \begin{tabular}{@{\extracolsep{9pt}}l ccc@{}}
      \toprule
              & \multicolumn{3}{c}{\textbf{HELOC}} \\
      \cline{2-4}

      Model & Baseline & Importance & KNN \\
      \midrule

     LogReg
        & 4.86$\pm$1.84&\textbf{3.95$\pm$1.69}&4.71$\pm$1.72\\

     MLP
        & 8.88$\pm$2.54&\textbf{8.34$\pm$2.58}&8.45$\pm$2.53\\

     GradBoost
         & 1.5$\pm$0.6&\textbf{1.49$\pm$0.58}&1.5$\pm$0.58\\

     SVC
         & 2.5$\pm$1.32&\textbf{2.01$\pm$1.14}&2.44$\pm$1.27\\

      \bottomrule
  \end{tabular}
\end{table}

\section{Conclusion}\label{sec:conclusion}
We explain credit application predictions obtained with black box models with counterfactuals. In case of positive prediction, we show how counterfactuals can be interpreted as a safety margin from the decision boundary. We propose two weighted strategies to generate counterfactuals: one derives weights from features importance, the other relies on nearest neighbours. Experiments on the HELOC loan applications dataset show that weights generated from feature importance lead to more compact counterfactuals, therefore offering more compact and intelligible explanations for end users.
Future work will focus on validating the effectiveness of our counterfactual explanations against human-grounded and application-grounded evaluation protocols (including the claim that smaller counterfactuals are indeed more interpretable). We will also experiment with weighting strategies that rely on model-specific feature importance, i.e. effect of feature perturbation on entropy of changes in predictions.

\bibliographystyle{unsrtnat}
\bibliography{main}

\end{document}